\documentclass[]{article}
\usepackage{authblk}
\pdfoutput=1
\usepackage[margin=1in]{geometry}

\usepackage{times}
\usepackage{helvet}
\usepackage{courier}
\usepackage{amsmath}
\usepackage[hyphens]{url}
\usepackage{graphicx}
\usepackage{graphicx}
\usepackage[]{todonotes}
\usepackage{booktabs} 
\usepackage{amsfonts}       
\usepackage{nicefrac}       
\usepackage{microtype}      
\usepackage{enumitem}
\usepackage{mathtools}
\usepackage{subfig}
\usepackage{xcolor}
\usepackage{mathrsfs}

\DeclareMathOperator*{\argmin}{arg\,min}

\newcommand{\MODELPARAM}{\theta} 
\newcommand{\DECISIONPARAM}{\omega} 
\newcommand{\DATAPOINT}{y} 
\newcommand{\DATASET}{\mathcal{D}}

\newcommand{\COVARIATE}{x} 
\newcommand{\POSTERIOR}{p(\MODELPARAM|\mathcal{D})} 
\newcommand{\APPROXIMATION}{q(\MODELPARAM)}
\newcommand{\MODEL}{p(\DATAPOINT,\MODELPARAM)} 
\newcommand{\PREDICTIVE}[1]{p_{#1}(\DATAPOINT)} 
\newcommand{\DECISIONMAKER}{f(p_{q}(\DATAPOINT),\DECISIONPARAM)} 
\newcommand{\DECISIONMAKERq}{f(p_{q}(\DATAPOINT),\DECISIONPARAM)} 
\newcommand{\DECISIONMAKERqn}{f(p_{q}(\DATAPOINT_n),\DECISIONPARAM)} 
\newcommand{\DECISION}{h} 
\newcommand{\ERISK}{\mathcal{ER}}
\newcommand{\RISK}[1]{\mathcal{R}_{#1}} 
\newcommand{\LOSS}{\ell(\DATAPOINT,\DECISION)}
\newcommand{\loss}{\ell}

\newcommand{\EXPECT}{\mathbb{E}}
\newcommand{\y}{\DATAPOINT}
\newcommand{\h}{\DECISION}
\newcommand{\x}{\COVARIATE}

\newcommand{\citep}{\cite}
\newcommand{\citet}{\cite}

\renewcommand{\todo}[1]{} 

\usepackage{hyperref}       
\hypersetup{
pdftitle={Correcting Predictions for Approximate Bayesian Inference},
pdfauthor={Tomasz Kusmierczyk, Joseph Sakaya, Arto Klami},
pdfkeywords={loss-calibrated variational inference, Bayesian inference, reasoning under uncertainty}
}

\begin{document}

\title{Correcting Predictions\\ for Approximate Bayesian Inference}

\author[]{Tomasz Ku\'smierczyk}
\author[]{Joseph Sakaya}
\author[]{Arto Klami}
\affil[]{
Helsinki Institute for Information Technology HIIT\\ 
Department of Computer Science, University of Helsinki, Finland\\ 
 \texttt{\{tomasz.kusmierczyk,joseph.sakaya,arto.klami\}@helsinki.fi}
 }

\maketitle

\begin{abstract}
\begin{quote}
Bayesian models quantify uncertainty and facilitate optimal decision-making in downstream applications. For most models, however, practitioners are forced to use approximate inference techniques that lead to sub-optimal decisions due to incorrect posterior predictive distributions. We present a novel approach that corrects for inaccuracies in posterior inference by altering the decision-making process. We train a separate model to make optimal decisions under the approximate posterior, combining interpretable Bayesian modeling with optimization of direct predictive accuracy in a principled fashion. The solution is generally applicable as a plug-in module for predictive decision-making for arbitrary probabilistic programs, irrespective of the posterior inference strategy. We demonstrate the approach empirically in several problems, confirming its potential.
\end{quote}
\end{abstract}

\section{Introduction}

Bayesian inference provides the fundamental basis for modeling
uncertainty. The posterior distribution $\POSTERIOR$ provides a complete summary of what is known about the parameters $\MODELPARAM$ of a model $p(\DATAPOINT, \MODELPARAM)$ given some observed data $\DATASET=\{(\COVARIATE_n, \DATAPOINT_n)\}_{n=1}^N$. In particular, the posterior is necessary and sufficient information for making optimal decisions under uncertainty \cite{berger}.

\begin{figure}[ht]
    \centering
    \includegraphics[width=0.55\columnwidth]{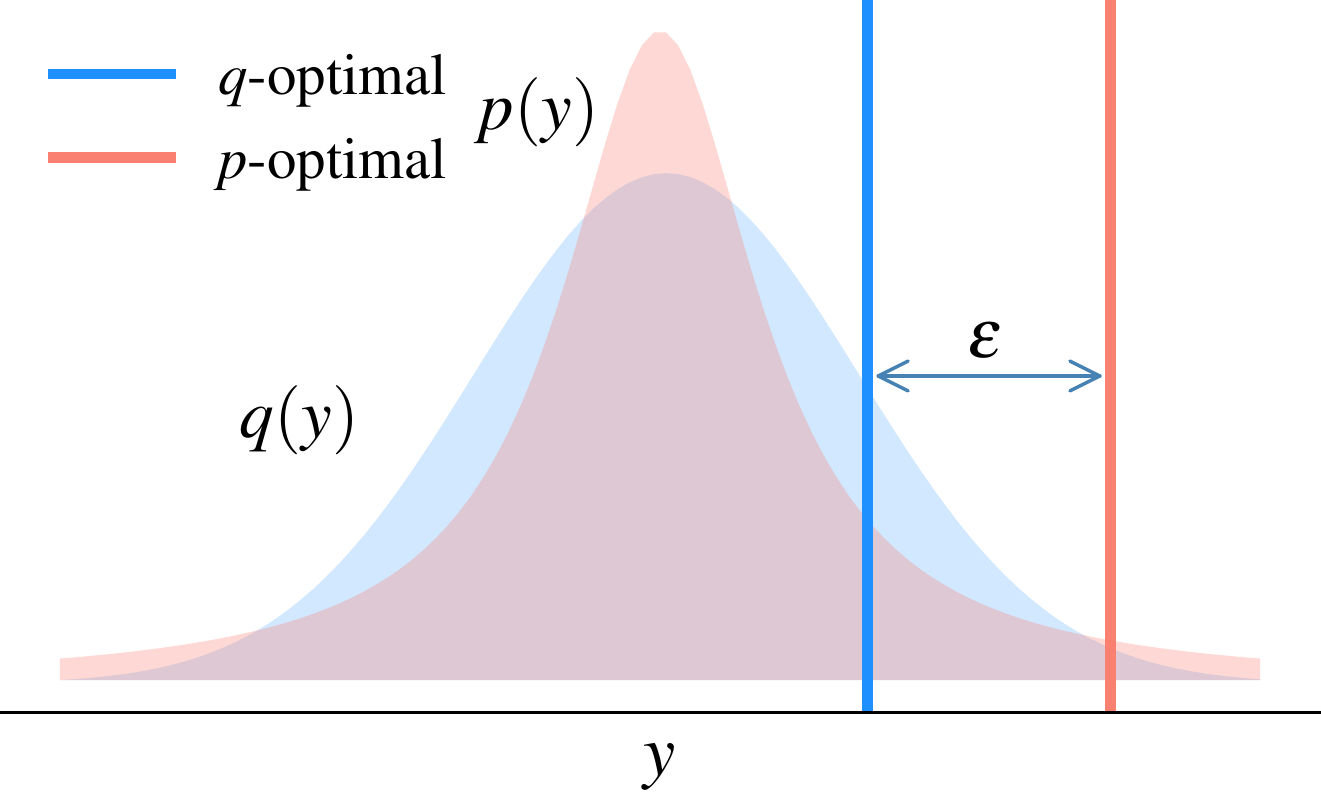}
    \includegraphics[width=0.55\columnwidth]{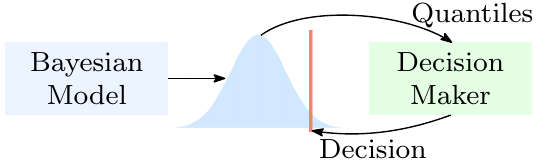}
    \caption{Bayesian decision theory gives a closed-form analytic rule for making optimal decisions for a given loss, based on the posterior predictive distribution (red).
    If the same decision rule is used for approximate predictive distribution (blue) based on approximate inference strategy, the numerical decision may be badly off as indicated by the gap $\epsilon$.
    Our approach learns to correct for this mistake, by replacing the theoretically optimal decision with parametric decision-making module -- a neural network -- that maps quantiles of the incorrect predictive distribution into optimal decisions.}
    \label{fig:toy}
\end{figure}

For predictive models $p(\DATAPOINT|\COVARIATE)$, such as regression or classification, the decisions correspond to committing to individual decision $\DECISION$ that minimizes the risk $\RISK{}$, the expectation of a loss $\LOSS$ over the uncertainty in the model parameters. For example, to minimize the squared loss we should commit to the mean of the predictive distribution $p(\DATAPOINT|\COVARIATE) = \int p(\DATAPOINT|\COVARIATE,\MODELPARAM) p(\MODELPARAM|\DATASET) d\MODELPARAM$, whereas for a loss that penalizes for overestimation of the true value the optimal decision is to report a suitably chosen lower quantile of the predictive distribution. For many losses the Bayes optimal decision is a simple statistic of the predictive distribution (Table~\ref{tab:losses}).

Unfortunately, the decisions obtained by minimizing the risk are optimal only if we have access to the true posterior $\POSTERIOR$. For most models we can only obtain an approximation $\APPROXIMATION \approx \POSTERIOR$. The almost universally adopted practice is to plug in the approximation in lieu of the posterior in the predictive distribution and proceed as though the result were the true predictive distribution. Even though this works relatively well for good approximations (e.g., well-converged Markov Chain Monte Carlo sampler), it may fail miserably for bad approximations. Furthermore, even for good approximations it would make sense to account for common biases; e.g., mean-field variational approximations are known to underestimate the posterior variation and hence also the variation in the predictive distribution \citep{blei2016VI}. Consequently, using decision rules optimal for the true posterior will have systematic bias as well.

Even though the problem persists in virtually all use of approximate Bayesian inference in decision-making problems, the literature addressing the issue is limited. The best examples build on the concept of \emph{loss-calibrated inference}. Originally proposed in the context of variational approximation by Lacoste--Julien et al.~\citet{lacoste2011approximate}, the core idea is to take into account the eventual decision problem (characterized by some loss) already during inference by altering the learning objective for approximate inference. Recently Cobb et al.~\citet{cobb2018loss} and Ku\'smierczyk et al.~\citet{lcvi} have presented practical algorithms building on this principle, but despite the elegant theory the procedure runs into several practical difficulties: It is computationally heavy, requires knowing the decision loss already during inference, and the empirical improvements are typically meagre. A similar approach has been proposed also for MCMC \citep{abbasnejad}, with similar drawbacks.

We propose an alternative, novel approach for making 
optimal
decisions based on a probabilistic model characterized by an approximate posterior. Instead of modifying the \emph{inference procedure} as in loss-calibrated inference, we modify the \emph{decision making phase} by replacing the analytic Bayes optimal decisions with a parametric function tuned to minimize the risk for the approximate posterior at hand. The proposed approach is conceptually simple, can be applied on top of any approximate posterior irrespective of the manner it was obtained, does not require re-computing the posterior approximation for new losses, and it is demonstrated empirically to clearly outperform the loss-calibrated inference approach \cite{lcvi} in some scenarios. When applied on the true posterior the procedure reverts back to the standard Bayes optimal decision.

Our key idea is to couple the Bayesian model with a decision-making module, in practice a neural network, which takes as input a characterization of the predictive distribution $p(\DATAPOINT|\COVARIATE)$ and outputs the decision $\DECISION$. It mimics optimal Bayesian decision rules by using the same input sufficient for optimal decisions, but is optimized to minimize a particular risk. The principle is illustrated in Figure~\ref{fig:toy}.

Training a flexible mapping from the predictive distribution to the decisions by direct optimization of the empirical risk can, naturally, overfit to the specific data collection. Furthermore, a flexible mapping may, under some conditions, overrule the underlying model, by making predictions not supported by the assumed model. We overcome these issues by presenting a generalized Bayesian inference strategy, building on \citet{bissiri}, over the decision-making modules. We provide a justified prior distribution that controls the decision-making module, by regularizing it towards the standard Bayes optimal decision, and demonstrate automatic prior specification using a light bootstrap procedure \citep{efron1994introduction}.

Our result is a plug-in tool that is applicable for every Bayesian model designed for predictive tasks. One can use any modeling framework or inference strategy, as long as we have access to samples from the predictive distribution. 
We demonstrate the approach for standard Bayesian models ranging from matrix factorization to sparse regression, and show improved decisions for poor approximations. 

\section{Background}

To facilitate understanding the rest of the paper, we briefly summarize Bayesian decision theory,  the concept of approximate posterior inference,  and the standard approach for incorporating approximate posteriors into Bayesian decision making.

\subsection{Bayesian Decision Theory}
\label{sec:bdt}

\begin{table}[tp!]
\centering
\caption{Example losses and their $p$-optimal decisions.}
{
\renewcommand{\arraystretch}{1.25}
\begin{tabular}{p{1.2cm} c c}
\toprule
{\sc Loss} & {\sc Expression} & {\sc Decision rule} \\
\midrule
squared & $(h-\y)^2$ &  $\text{mean}_p[y]$ \\
absolute & $|h-\y|$ &  $\text{median}_p[y]$ \\
imbalanced absolute 
&
{
$\left\{
\begin{array}{ll}
 a \cdot |\h-\y| & \y \geq \h \\
 b \cdot |\h-\y| & \y < \h
 \end{array}
\right.$
}
& $\frac{a}{a+b}$-percentile$_p[y]$ \\
tilted &
{
$\left\{
\begin{array}{ll}
 t \cdot |\h-\y| & \y \geq \h \\
 (1-t) \cdot |\h-\y| & \y < \h
 \end{array}
\right.$
}
& $t$-percentile$_p[y]$ \\
\bottomrule
\end{tabular}
}
\label{tab:losses}
\end{table}

Bayesian decision theory~\cite{berger} defines a rigorous framework for decision-making under uncertainty. Given a loss function $\loss(\MODELPARAM, h)$ and a posterior distribution $p(\MODELPARAM|\DATASET )$ of a parametric model conditioned on the dataset $\DATASET$, the Bayes optimal decision $h_p$ minimizes expected posterior loss 
$
\mathcal{R}_p(h) = \int  \loss(\MODELPARAM, h)p(\MODELPARAM|\mathcal{D})\ d\MODELPARAM.
$

In supervised settings, decisions are made 
for pairs $(x,y)$, and
the risk 
$\mathcal{R} = \EXPECT_{(x,y) \sim p_\DATASET} \ell(y,h(x))$ is an expectation over
unknown data generating distribution $p_\DATASET$. 
For setups where we do not explicitly model $p(\x)$ but do assume a model for $p(y|x)$, we can
define conditional risk for $x$ using 
\begin{equation}
\RISK{p}(h|x) = \int  \LOSS p(y|\DATASET, x) dy.
\label{eq:risk}
\end{equation}
where
$p(\DATAPOINT|\DATASET, x) = \int p(y|\theta,x) p(\theta | \DATASET) d\theta$.
The optimal decisions for individual data points are then  $h_p =$ $\argmin_{h\in\mathcal{H}}$ $\RISK{p}(h|x)$, denoted as \emph{$p$-optimal} 
to indicate they are optimal for the true predictive distribution. 
Similarly, we call \eqref{eq:risk} the \emph{$p$-risk} or simply risk.
Table~\ref{tab:losses} presents $p$-optimal decisions for several example losses.

In practice the risk needs to be approximated  by \emph{empirical risk} $\ERISK = \sum_{n =1} ^{N} \loss(y_n, h(x_n))$, based on some data collection, 
where the distribution of $x$ and $y$ is purely empirical.
More detailed discussion of Bayesian decision theory for supervised learning
can be found in \citet{lacoste2011approximate}.

\subsection{Approximate Inference}
Typically, the posterior distribution $p(\MODELPARAM|\DATASET)$ is analytically intractable, and we need to resort to approximate inference techniques to replace it with a computationally tractable proxy $q(\MODELPARAM)$. A wide array of  inference tools are available with different trade-offs between computational time and efficiency. Monte Carlo Markov Chain (MCMC) methods 
approximate $p(\MODELPARAM|\DATASET)$ with samples from the posterior, with state-of-the-art algorithms using gradient-based information to improve convergence and speed \citep{nuts}.
Distributional approximations, in turn, assume a parametric approximation family $q(\MODELPARAM; \alpha)$ and optimize for its parameters $\alpha$ to minimize some discrepancy measure between the approximation and the true posterior. This can be done in multitude of ways, the most common strategies being variational approximation \citep{blei2016VI} and expectation propagation \citep{gelman2017expectation,minkaep}. Distributional approximations often have computational advantage over MCMC, but are biased or inaccurate especially when using poor approximation family \citep{Yao2018diditwork}. For example, variational approximation minimizes a Kullback-Leibler divergence between the approximation and the true posterior, and hence underestimates posterior variation \citep{blei2016VI}.

Probabilistic programming tools, such as Stan \citep{carpenter2017stan} and Edward \citep{edward} have recently made practical Bayesian modeling easier, by coupling automatic inference engines based on Hamiltonian Monte Carlo \citep{nuts} or gradient-based variational approximation \citep{dsvi,kucukelbir2017automatic} with easy model specification language. The typical mode of operation today is to specify the model in such a language, and the practitioner may not even care about the approximation strategy being used. This has made Bayesian modeling possible for wider audience, which implies more research is needed on how the models are being used in downstream applications.

\subsection{Decision-making for Approximate Inference}
\label{sec:approximate_decision_making}

We often do not have access to the true posterior  $p(\theta | \mathcal{D})$ to evaluate  the $p$-risk in Eq.~\ref{eq:risk}.
We instead use the approximation $q(\MODELPARAM)$ as a proxy for the posterior. 
Following the nomenclature of Lacoste--Julien et al.~\citet{lacoste2011approximate},
we define the \emph{$q$-posterior predictive distribution} $p_q(\DATAPOINT|\DATASET, x) = \int p(y|\theta,x) q(\theta) \ d\theta$ and the
\emph{$q$-risk}
\begin{equation*}
\RISK{q}(h|x) = \int  \LOSS  p_q(y|\mathcal{D}, x) dy.
\end{equation*}
The \emph{$q$-optimal} decisions $h_q = \argmin_{h\in\mathcal{H}} \RISK{q}(h|x)$ minimize the $q$-risk. 
Despite the name, the $q$-optimal decisions 
are not optimal with respect to the $p$-risk.
Instead, we typically have
$
\RISK{p}(h_q|x) > \RISK{p}(h_p|x)
$.
Section~\ref{sec:pdt} describes how the discrepancy between $h_q$ and $h_p$ can be mitigated, when only having access to the approximation $q{(\MODELPARAM)}$. 

\section{Predictive Decision Theory}
\label{sec:pdt}

\todo{Should we here all the time use $p(y|x)$ instead of just $p(y)$?}
Our goal is to make good -- possibly even optimal -- decisions, as measured by the risk $\RISK{}$ under a probabilistic model $\MODEL$, when only having access to an approximation $\APPROXIMATION$ of the true posterior $\POSTERIOR$. We do this by introducing a parametric decision-making module $\DECISIONMAKER$, a function that takes as input a characterization of the predictive distribution $\PREDICTIVE{q}$ under the approximate posterior and produces as an output the decision $\DECISION$. This module can be applied on top of arbitrary probabilistic models.

In the following, we describe the necessary technical elements required for implementing the proposed Predictive Decision Theory (PDT). We first formulate the problem as generalized Bayesian inference over the decision-makers in Section~\ref{sec:decisionbeliefs}, and then explain how practical decision-making modules can be implemented in Section~\ref{sec:decisionmaker}.

\subsection{Decision Belief Distributions}
\label{sec:decisionbeliefs}

To provide justified uncertainty quantification for the decisions, 
we build on the generalized Bayesian inference framework by Bissiri et al.~\citet{bissiri}. They present a coherent procedure for updating beliefs for scenarios where the parameter of interest is connected to observations via arbitrary loss functions, instead of likelihood functions. Their key result is that (using generic notation to avoid confusion with the symbols used in main derivations of our results)
\begin{align*}
p(\eta|y) \propto e^{-\loss(\eta, y)} p(\eta)^{\lambda_1}
\label{eq:beliefupdate}
\end{align*}
is a valid posterior distribution for any loss $\loss(\eta,y)$ that depends on the parameters $\eta$ and the observation $y$, and any constant ${\lambda_1} > 0$ allowing for calibration of relative impact of the prior and the loss.
In other words, we can use exponentiated negative losses in place of likelihoods and still characterize the uncertainty on $\eta$ in a coherent manner. 

The framework provides direct basis for decision-problems involving the model parameters $\MODELPARAM$ itself. We use it for predictive problems instead, for problems characterized by loss $\loss(h,y)$ defined in terms of predicted quantities. 
The decision is made based on the predictive distribution $p(y)$ that is obtained via marginalization of model parameters $\MODELPARAM$.
One could think of plugging in the predictive distribution along with the procedure for obtaining $h$ in place of loss, but that would make model the impossible to train because of the marginalization required over $\MODELPARAM$.  
Instead, we consider \emph{decision belief distributions}
\begin{align*}
p(h|y) \propto e^{-\loss(h, y)} p(h)^{\lambda_1}
\end{align*}
and tie decisions for different $\DATAPOINT$  with parameteric \emph{decision-making module} (DM) $h = \DECISIONMAKER$.
DM makes decisions on the observed data $\DATAPOINT$, is parameterized by $\DECISIONPARAM$, and takes as input the predictive distribution $p(y)$
(or an approximation $\PREDICTIVE{q}$) from already trained model. 
This allows us to define the Bayesian update rule for belief distribution of a decision-making module
\begin{equation}
p(\DECISIONPARAM|y) \propto e^{ -\loss'(\DECISIONPARAM, \DATAPOINT)} p(\DECISIONPARAM)^{\lambda_1}
\label{eq:dm_updates},
\end{equation}
where 
$\loss'(\DECISIONPARAM, \DATAPOINT):=\loss(\DECISIONMAKER, \DATAPOINT)$ and
$p(\DECISIONPARAM)$ is some prior defined over the parameters of the DM. The constant ${\lambda_1}>0$ allows tuning the compromise, but importantly all values result in valid posterior. 

\subsection{Decision Maker}
\label{sec:decisionmaker}

We replace the standard decision-making procedure explained in Section~\ref{sec:approximate_decision_making} with parametric function (decision maker) $\DECISIONMAKER$, for which posterior inference is characterized by Eq.~\eqref{eq:dm_updates}. 
For a collection of $N$ data instances the log-posterior
is
\begin{equation}
\log p(\DECISIONPARAM|\mathcal{D}) = -{\sum_{n=1}^N \loss'(\DECISIONPARAM, \DATAPOINT_n)} + \lambda_1 \log p(\DECISIONPARAM) + C' \propto  -\underbrace{\frac{1}{N}\sum_{n=1}^N \loss'(\DECISIONPARAM, \DATAPOINT_n)}_{\ERISK} + \lambda \log p(\DECISIONPARAM) + C,
\label{eq:MAP}
\end{equation} 
where $C$ is a normalization constant that has no effect on inference and $\lambda := \frac{\lambda_1}{N}$.
Empirical risk $\ERISK$ appears naturally as a part of the objective. 
Suitable priors, however, help us to address the issue of generalization, 
to make sure the decisions are good for the true
risk instead of just the empirical one.
The formulation defines the basis for 
full posterior inference,
but we focus on finding a single good decision-maker and hence resort to
 MAP estimation of Eq.~\eqref{eq:MAP}
 that is less computationally expensive.

\subsubsection{Regularization with $q$-optimal Decisions}

We start by specifying the prior $p(\DECISIONPARAM)$ used for controlling the flexibility of the DM, by building on the assumption that $\APPROXIMATION$ is reasonably good approximation of $\POSTERIOR$. When they are identical the $q$-optimal decision $\DECISION_q$ corresponds exactly to the $p$-optimal decision $\DECISION_p$, and for small approximation errors we would still expect to find the optimal decision in the neighborhood of the theoretical optimum. 

Building on the above principle we define an implicit prior.
Instead of placing a prior on the parameters $\DECISIONPARAM$ directly, we place a prior on the decisions $\DECISION=\DECISIONMAKERq$, which induces a prior on $\DECISIONPARAM$. For the decisions we assume the prior
\begin{equation}
    \DECISION^n  \sim \mathcal{N}(h^n_q, 1),
    \label{eq:h_simple_prior}
\end{equation}
where $h^n_q$ is $q$-optimal decision for $n$th data point and the variance is set to unit value as it can be subsumed to the $\lambda$ parameter in Eq.~\eqref{eq:MAP}. Small $\lambda$
allows DM to deviate arbitrary far from $h_q$, whereas large $\lambda$ forces DM to emulate $q$-optimal decisions.
A good value can be selected, e.g., by cross-validation.

A more flexible prior is obtained by allowing 
point-specific variation with 
$$
\DECISION^n  \sim \mathcal{N}(\mu_h^n, \sigma^n_h).
$$
This prior requires an automatic procedure for setting $\mu_h^n$ and $\sigma_h^n$, which we do based on the expected variation caused by the underlying distribution $p(\COVARIATE)$. The decisions are made conditional on the covariates and the total risk averages over the data generating distribution
$p_{\DATASET} (x,y) \approx p(y|x) p(x)$, as briefly explained in Section~\ref{sec:bdt}. The natural variation in samples produced by the generating distribution provides a reasonable basis for estimating the allowed variation in decisions as well, and to estimate this we apply a bootstrap procedure \citep{efron1994introduction}.

With bootstrapping, we draw several new datasets $\DATASET_l$ ($l = 1 \dots L$), using $L=5$ in our empirical experiments.
For each dataset, we obtain the posterior approximation $\APPROXIMATION$ and the $q$-optimal decisions $h_{ql}$, and
estimate the parameters for each data point using
$\hat \mu_h^n = \frac{1}{L} \sum_{l=1}^L h^n_{ql}$,
$\hat \sigma^n_h = \sqrt{\frac{1}{L} \sum_{l=1}^L (h^n_{ql} - \hat \mu_h^n)^2 }$.
Finally, under independence assumption on $h^n$, we can define the prior for $\DECISIONPARAM$ as 
\begin{equation}
\log p(\DECISIONPARAM) = \sum_{n=1}^N \log N(\DECISIONMAKERqn |  \mu^n_h, \sigma^n_h)
\label{eq:simple_prior},
\end{equation}
where for a simpler case $(\mu^n_h, \sigma^n_h) := (h_q^n, 1)$ and for bootstrap-based prior
$(\mu^n_h, \sigma^n_h) := (\hat \mu^n_h, \hat \sigma^n_h)$. The overall strength of the prior is in both cases controlled by $\lambda$.

\subsubsection{Representation of the Predictive Distribution}
\label{sec:representation_predictive}

Bayesian decision theory states that the predictive distribution $\PREDICTIVE{}$ is necessary and sufficient information for making optimal decisions. We retain this assumption when switching to parametric decision-makers, by using the predictive distribution as the  sole input for the decision-maker. However, it is rarely available as analytic expression.

A practical and model-independent representation is obtained by sampling a collection of $S$ data points $\DATAPOINT_s$ from the model $\MODEL$ under the approximate posterior $\APPROXIMATION$. Such a collection can be summarized with suitable finite statistic that can then be passed as input for the decision-maker; we use $B$ evenly-spaced empirical quantiles, but for example a histogram would work as well. This representation only requires the ability to sample from the predictive distribution, covering essentially all probabilistic models of interest. It could even generalize for simulator-based models that lack closed-form expression for likelihood but still enable sampling from the model \cite{Gutmann16}.

The parameters $B$ and $S$ influence the richness and accuracy of the input representation, but as illustrated empirically in Section~\ref{sec:lcvicomparison} the procedure is not very sensitive to the choices. Since computing the representation is cheap we can safely use large $S$, and small $B$ is enough because the decision-makers can implicitly interpolate between the available quantiles if needing more granularity.

\subsubsection{Parametric Decision Makers}

Most Bayesian optimal decisions are simple summary statistics of the predictive distributions (Table~\ref{tab:losses}), and simplified illustrations of the procedure like Figure~\ref{fig:toy} intuitively suggest that it may be enough to modify the specific threshold, e.g., by lowering or increasing the quantile for tilted loss to account for the discrepancy between the predictive distributions. Such a decision-maker is easy to implement and has limited room for overfitting, but we will later show empirically (Section~\ref{sec:lcvicomparison}) that it is not sufficient for notably improving the predictions. Instead, we need more flexible functions.

A natural choice for a flexible model-independent decision maker is to use a neural network, interpreted as arbitrary mapping from the predictive distribution $\PREDICTIVE{}$ to the decision $\DECISION$. The decision-maker is parameterized by $\DECISIONPARAM$, which denotes collectively the set of all network weights. Such modules are easy to implement in modern machine learning platforms, allowing for flexible choice of network architectures. In our experiments, we use  a simple feed-forward network with 3 hidden layers (with  20, 20 and 10 nodes) with ReLU activation and Adam optimizer with learning rate $= 0.01$, but note that the specific network details are not important. Instead, we simply need a network that is able to relatively flexibly convert a fixed-dimensional quantile representation of the predictive distribution into a scalar decision.

\begin{figure*}[tp!]
    \centering
    \includegraphics[width=0.33\textwidth]{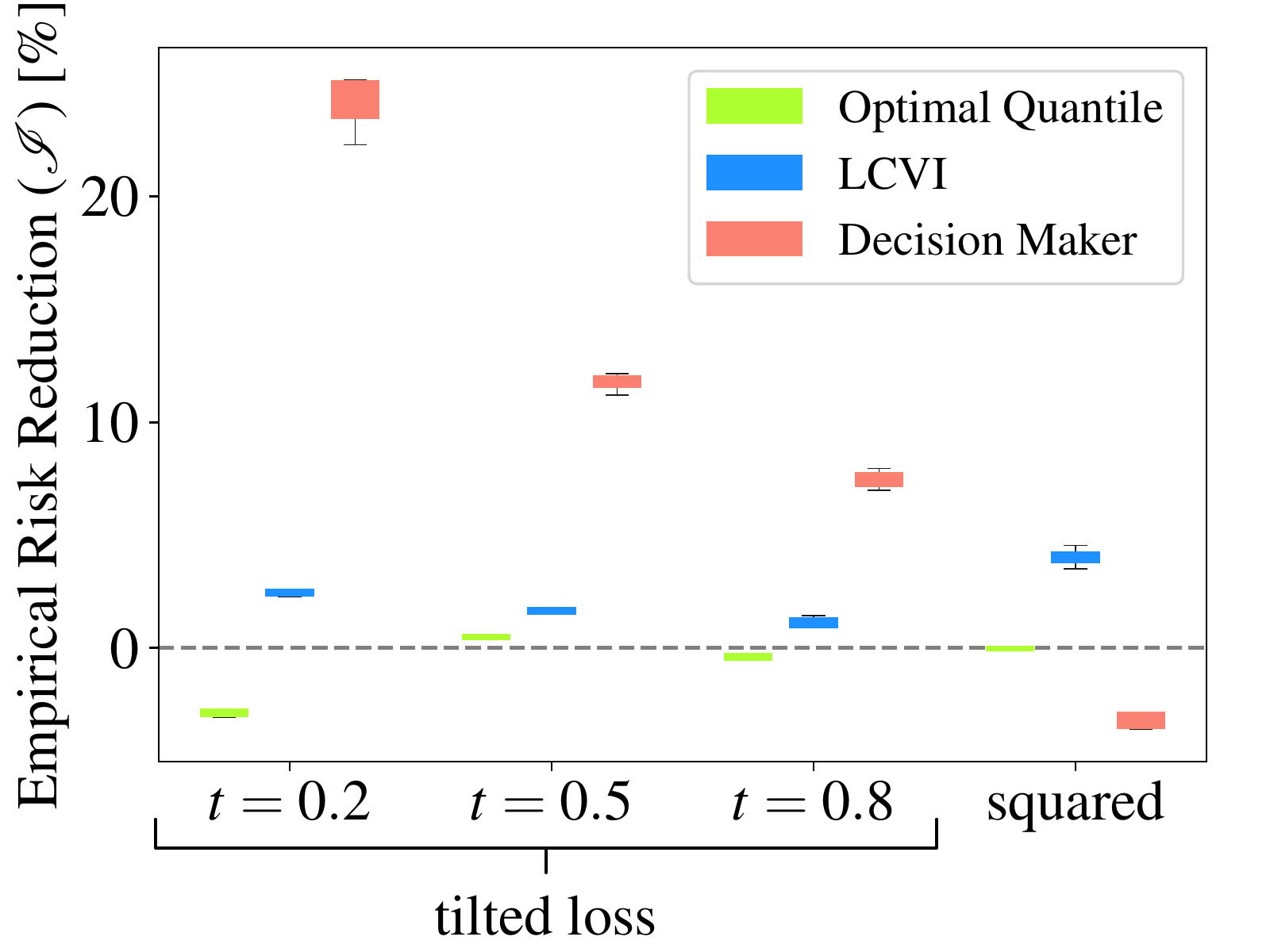}
    \includegraphics[width=0.33\textwidth]{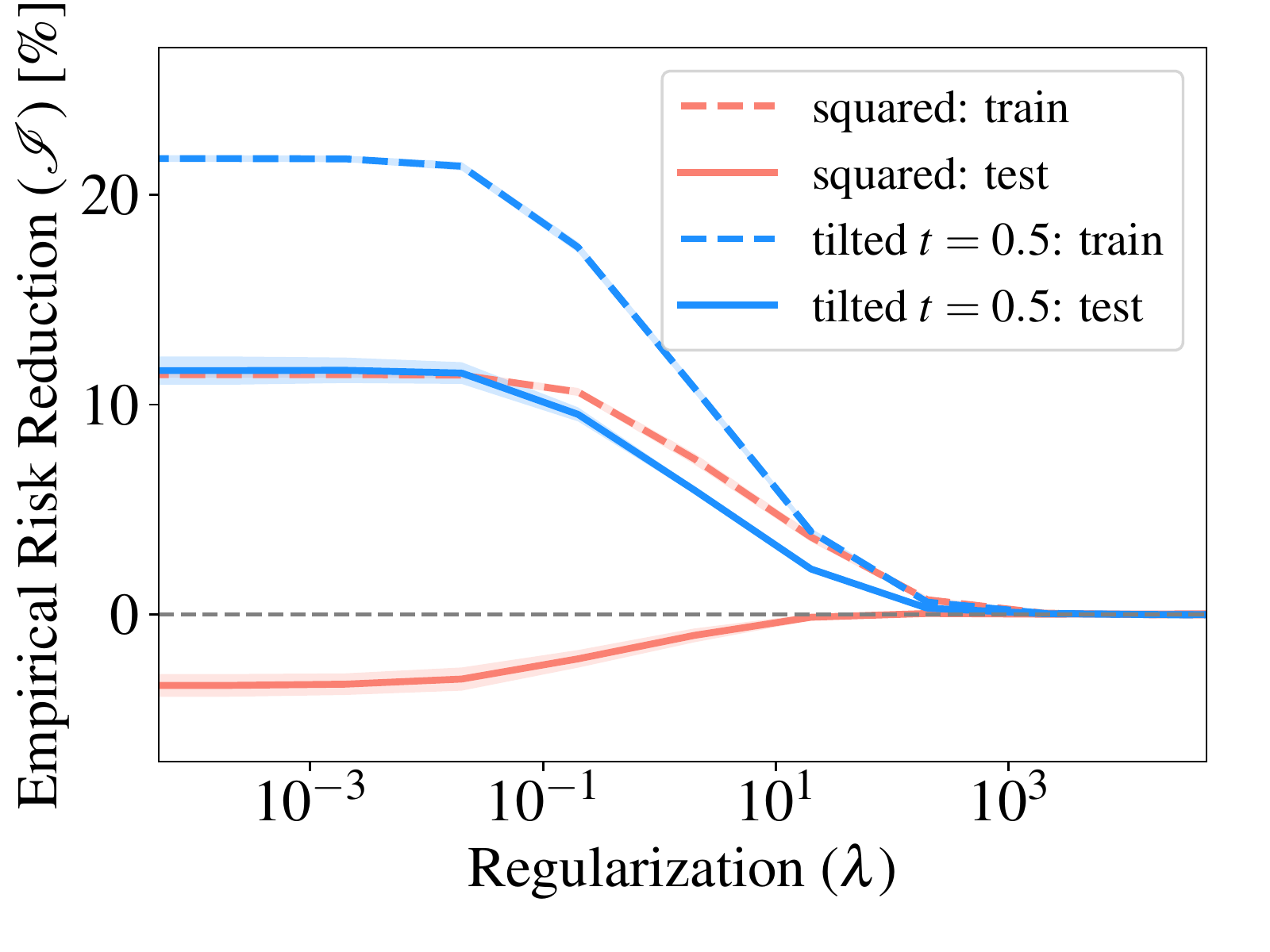}
    \includegraphics[width=0.33\textwidth]{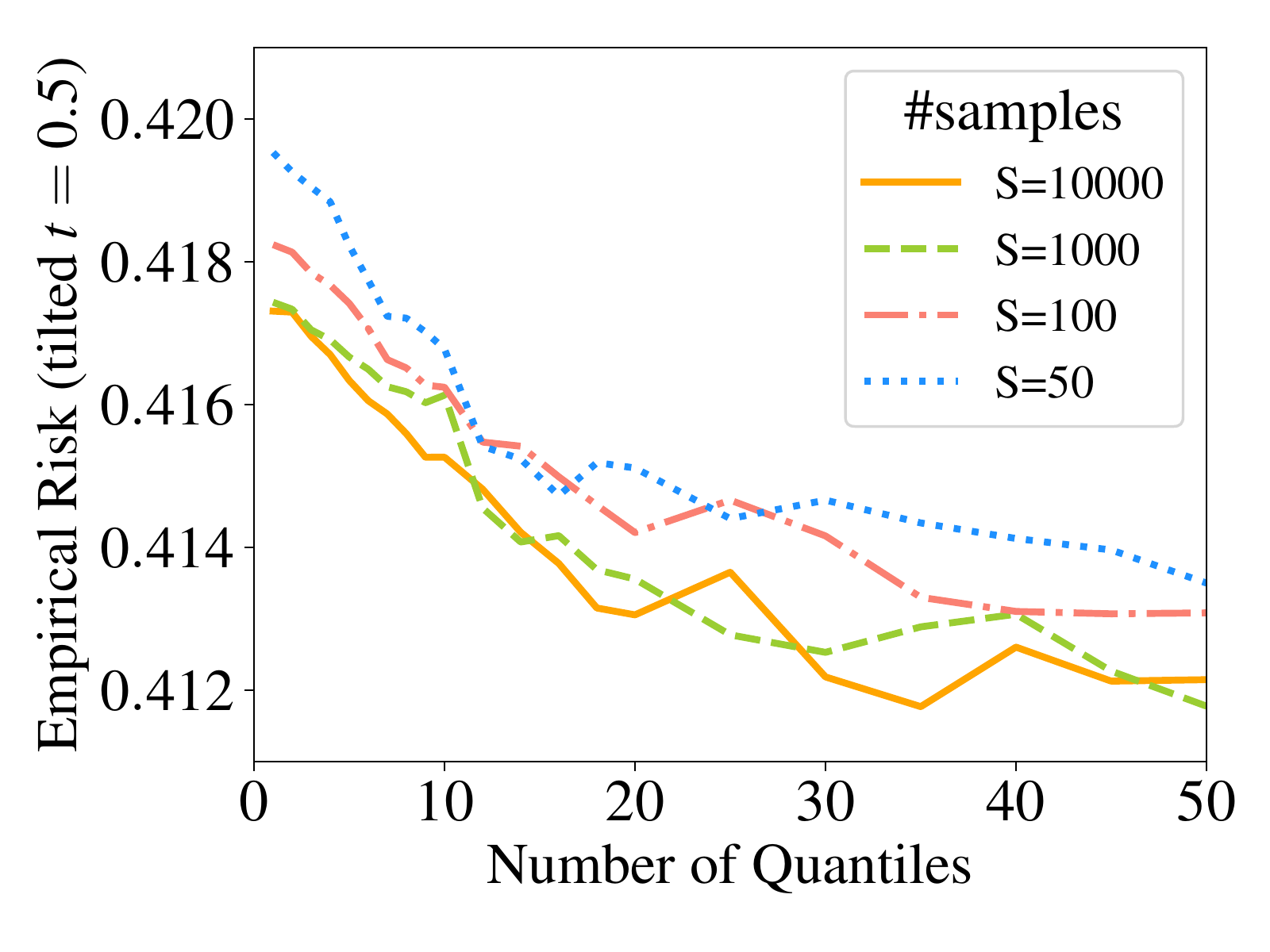}
    \caption{Matrix Factorization of \textit{last.fm} data: 
(Left:) Predictive performance on test data, showing the proposed method clearly outperforms LCVI for tilted losses. 
(Middle:) Regularization effects on test and training data for squared and tilted ($t=0.5$) loss, showing how the regularization completely removes overfitting.
(Right:) Effects of number of samples and predictive distribution representation (number of quantiles) on test data performance for tilted ($t=0.5$) loss.}
    \label{fig:mf}
\end{figure*}

\section{Related Work}
\label{sec:related_work}
Our work lies in the intersection of Bayesian modeling and learning theory. The general principle of minimizing the regularized risk for making decisions that generalize for the data-generating distribution is the cornerstone of standard learning theory (see, e.g., \citet{hastietibshirani}). The current work does not utilize or require results in recent learning theory literature, but we note that there may be interesting connections useful for future work. Our method also loosely relates to work on quantile regression \cite{quantile_regression1} and the training of decision-makers to generate prediction intervals \cite{quantile_regression2}. However, we build on existing Bayesian models and merely use a neural network to correct for approximation error, whereas they use black-box methods for the predictive distribution itself.

The most closely related work is on direct loss-calibration of posterior approximations \citep{lacoste2011approximate,abbasnejad}. We propose an alternative for that approach and will empirically compare the proposed solution against the latest loss-calibration method \citep{lcvi}, and hence describe here their approach briefly.

The core idea of loss-calibration is to replace the original learning objective, in case of variational approximation, a lower bound to the marginal likelihood $\log p(\DATASET)$, with an augmented objective that involves a separate term accounting for the loss
\begin{equation*}
  \underbrace{\mathbb{E}_{q(\MODELPARAM ; \alpha)}
\left[\log p(\DATASET, \MODELPARAM)  - \log q(\MODELPARAM; \alpha) \right] }_{\text{lower bound to} \log p(\DATASET)} + 
\underbrace{ \mathbb{E}_{q(\MODELPARAM ; \alpha)} \left[ \log u(\MODELPARAM, h) \right]}_{\text{utility term}}.
\end{equation*}
The function $u(\MODELPARAM, h)$ is the loss $\ell(\MODELPARAM, h)$ transformed to utility \cite{berger}. This augmented objective is maximized, typically with an alternating algorithm, with respect to both the parameters $\alpha$ of the approximation and the decisions $\DECISION{}$. The decisions influence the approximation and the approximation influence the decisions, and hence the optimization is tightly coupled. This implies that the procedure needs to be carried out separately for all losses of interest and the algorithmic details need to be derived for each approximation strategy separately. Our approach is applied on top of existing approximation and hence does not suffer from these limitations: Once the approximation is available, we can easily optimize for any loss without needing to re-iterate the inference.

\section{Experiments}

To illustrate and validate the procedure we conduct a series of experiments, using variational approximation as $\APPROXIMATION$. We first compare the method against the alternative of calibrating the posterior inference to account for the loss 
for a matrix factorization model, and then demonstrate improved decisions for a sparse regression model and a multilevel model for cases with approximations. We also study practical decisions regarding the input representation and regularization. 

\subsection{Predictive Performance}
\label{sec:lcvicomparison}

We compare our approach to Loss-calibrated Variational Inference (LCVI) by Ku\'smierczyk et al.~\citet{lcvi}, the closest alternative to our method, using the same simplified probabilistic matrix factorization model 
\[
    Y \sim \mathcal{N}(ZW, \sigma_y), \quad
    W_{ik} \sim \mathcal{N}(0, \sigma_W), \quad Z_{kj} \sim \mathcal{N}(0, \sigma_z)
\]
they used for demonstrating loss-calibration improves predictions over standard variational inference on a subset of \textit{last.fm} data \citep{lastfm}. We use their code\footnote{https://github.com/tkusmierczyk/lcvi} to precisely reproduce the experiment, matching all of their modeling and approximation choices ($K=20$, $\sigma=10$, mean-field Gaussian approximation, and log-transformation for the count data).

Figure~\ref{fig:mf}~(left) compares relative reduction of empirical risk ($\mathscr{I} = \frac{\ERISK_\text{VI}-\ERISK}{\ERISK_\text{VI}}$, where
$\ERISK_\text{VI}$ is the risk of $q$-optimal decision on standard approximation)
on the test data for the same four losses they used. For tilted losses the proposed approach using neural network as decision-maker dramatically outperforms LCVI, reaching $7-24\%$ risk reduction depending on the $t$ parameter of the loss. This demonstrates the proposed approach has clear practical value over the alternative of fine-tuning the approximation itself. The simpler decision-maker that only optimizes for the quantile is unable to improve the decisions, demonstrating that it is preferable to learn a more flexible mapping.

For squared loss even the neural decision-maker performs poorly, due to overfitting. Figure~\ref{fig:mf}~(middle) illustrates how regularizing the decision-maker towards the $q$-optimal decisions using the prior \eqref{eq:simple_prior} elegantly removes the overfitting. The method never achieves risk reduction, but instead converges to the standard $q$-optimal estimator. Consequently, there are scenarios for which the computationally more expensive LCVI is able to improve the predictions, whereas the proposed method is not.

\subsection{Posterior Representation}

In practical computation the posterior predictive distribution is presented using empirical quantiles of samples drawn from the predictive distribution (Section~\ref{sec:representation_predictive}). The choice of the sample size $S$ and the number of quantiles $B$ naturally influences the representation accuracy. We study the effect of these parameters on the same data and model as above.

Figure~\ref{fig:mf} (right) compares test data empirical risk averaged of 10 different random initializations for various representations. The main observations are that it is beneficial to use large sample size, and since the computational cost of drawing the samples is negligible compared to other stages of the procedure (fitting the approximation, training the decision-maker) we use $S=1000$ in other experiments. The risk improves also when increasing the representation size, and for the other experiments we use $B=20$ quantiles. However, it is worth noting that already very few quantiles are sufficient for improving the risk compared to $q$-optimal decisions ($q$-risk $= 0.46$).

\subsection{Handling VI Failures} 
\label{sec:experiments_vi_failures}

The main use for the method is in correcting for inaccurate posterior approximations, and hence we conduct two experiments on two separate regression models to validate this. At the same time, we illustrate the bootstrap-based regularization strategy.

\subsubsection{Multilevel Models with Poor Approximation}

\begin{figure}[t]
    \centering
    \includegraphics[width=0.35\textwidth]{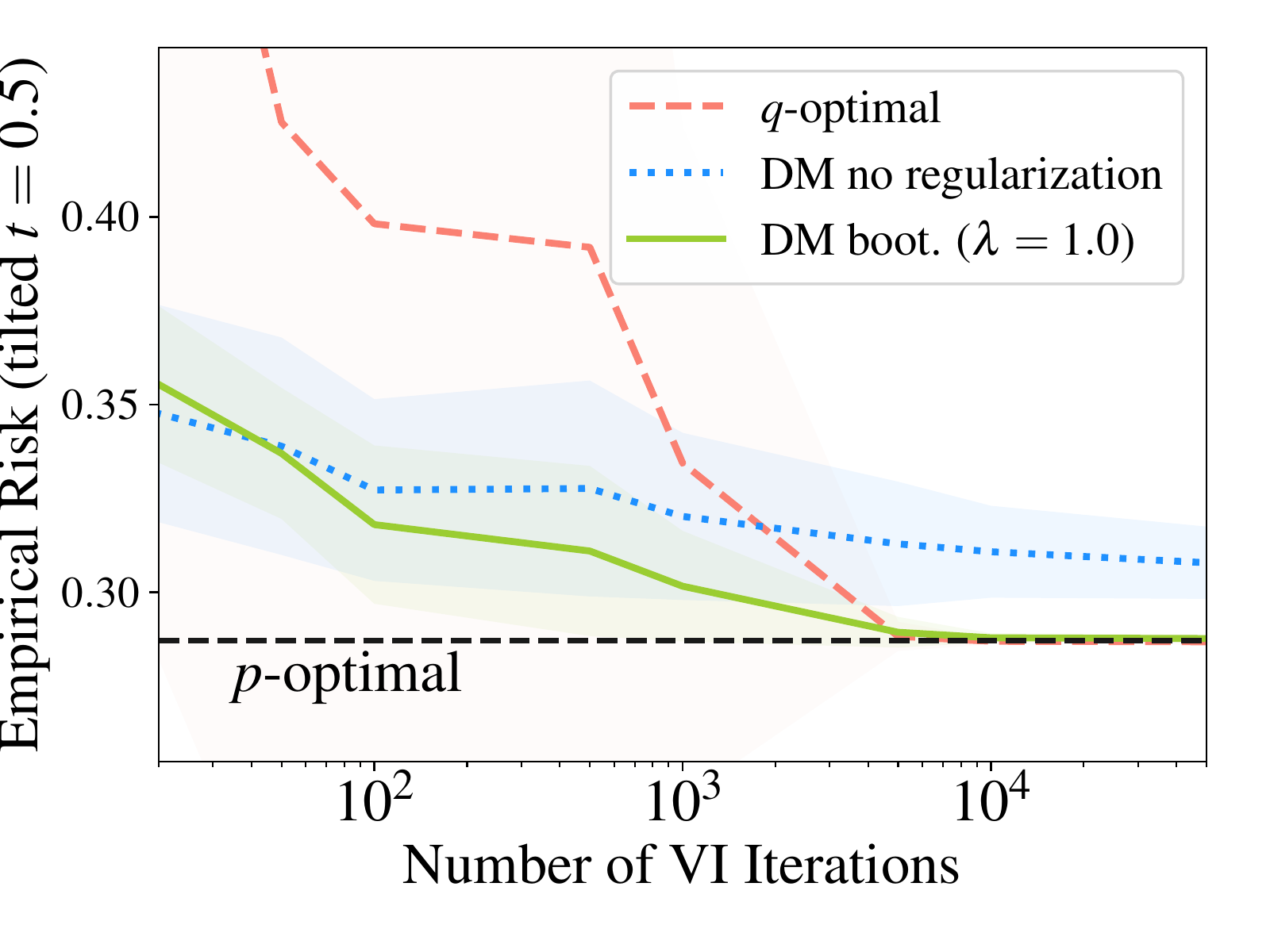}
    \includegraphics[width=0.35\textwidth]{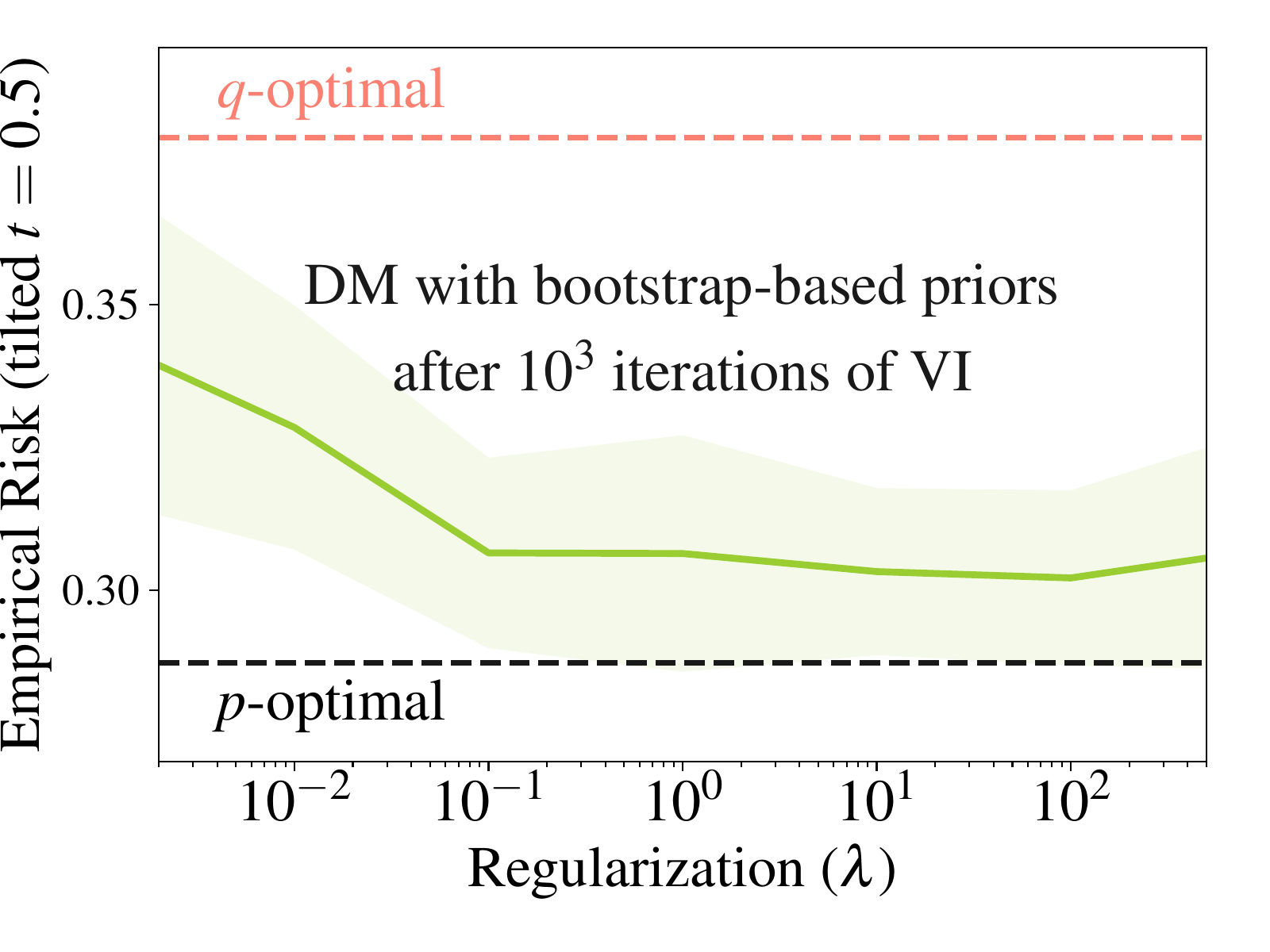}
    \caption{(Top:) Risk for tilted loss ($t=0.5$) on test subset of \textit{radon} data. The proposed method with regularization (solid green) is able to significantly reduce the risk over the $q$-optimal baseline (dashed red) for poor approximations. (Bottom:) The method is very insensitive to the regularization parameter $\lambda$; all choices improve over the $q$-optimal baseline.}
    \label{fig:radon}
\end{figure}

Since variational approximations are trained with iterative algorithms, we can construct a controlled experiment by terminating the inference algorithm early and varying the termination point. In general, approximations trained for shorter time are further away from the final approximation and the true posterior.

We use the \textit{radon} data and the multi-level model
\[
y_i = \alpha_{j[i]} + \beta x_i + \epsilon_i, \quad
\alpha_j = \gamma_0 + \gamma_1 u_j + \zeta_j, \quad
\zeta_j \sim \mathcal{N}(0, \sigma^2_\alpha)
\]
for modeling it \citep{gelman2006data}. The model is implemented using the publicly available Stan code\footnote{http://mc-stan.org/users/documentation/case-studies/radon.html}, using the Minnesota subset of the data and also otherwise conforming to their details. The inference is carried out using automatic differentiation variational inference \citep{kucukelbir2017automatic}, and we split the data randomly into equally sized training and test set. 

Figure~\ref{fig:radon} (top) compares performance of different decision-making strategies (vertical axis) w.r.t quality of the approximation fit controlled by number of training iterations (horizontal axis). Hamiltonian Monte Carlo \citep{nuts} provides here sufficiently accurate posterior to act as the baseline, and we see that variational approximation converges sufficiently close to the true posterior around $10^4$ iterations, reaching the $p$-risk. The proposed method without regularization achieves relatively good risk already for very poor posterior approximations, but never converges to the optimal decision due to overfitting. The variant regularized with the bootstrap-based prior in Eq.~\eqref{eq:simple_prior} achieves the best of both worlds, reducing the risk for poor approximations but converging to the $p$-optimal decisions when the approximation becomes good. Figure~\ref{fig:radon} (bottom) plots the risk at $10^3$ iterations (still incorrect approximation) as a function of the regularization parameter $\lambda$, showing that the method is robust to the choice of $\lambda$; all values within the range improve compared to the $q$-optimal baseline, and for $\lambda \in [10^{-1},10^2]$ the risk is virtually identical.

\subsubsection{Sparse Models with Failure of Convergence}

\begin{figure}[t]
    \centering
    \includegraphics[width=0.35\textwidth]{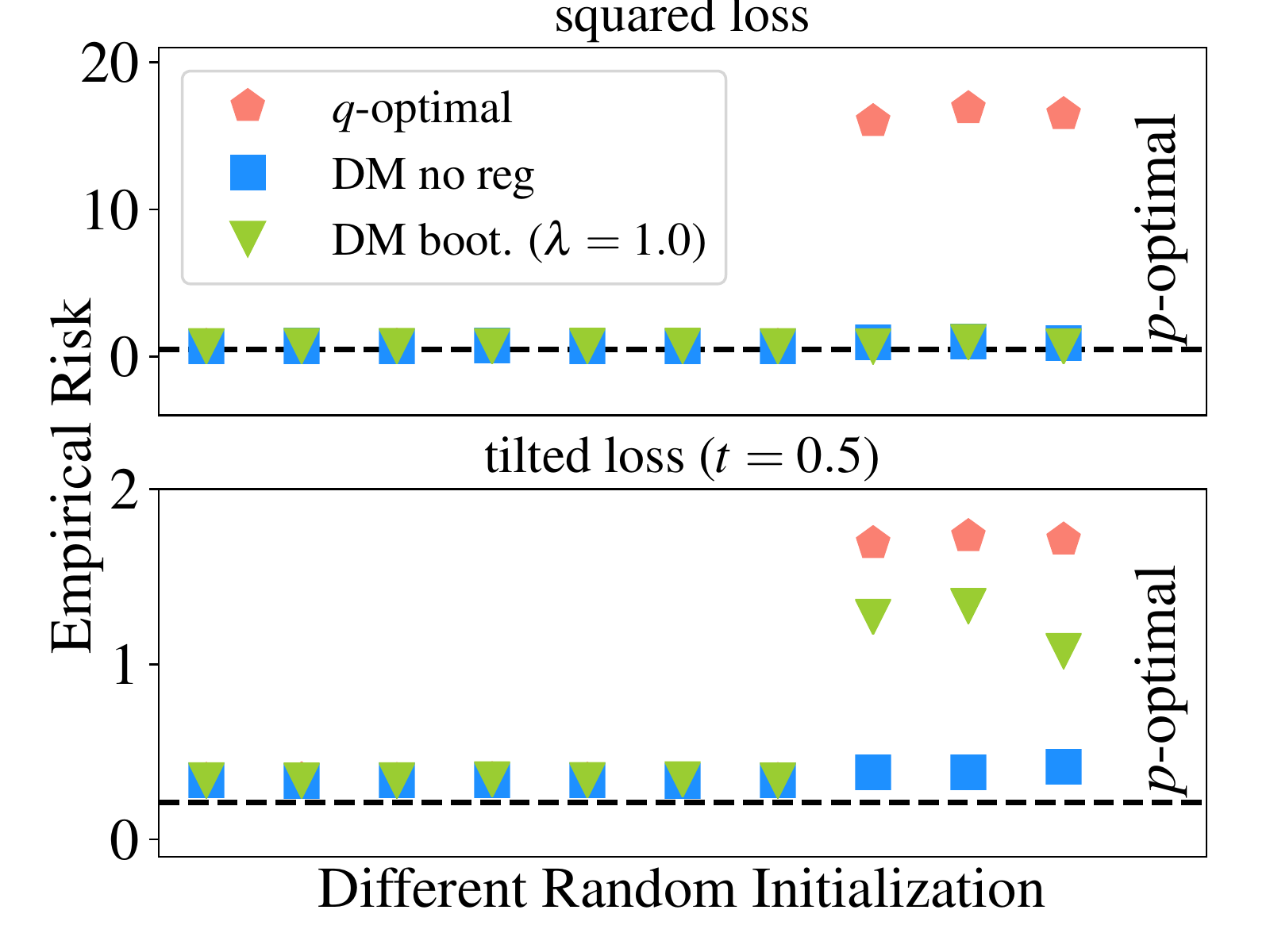}
    \caption{Risk for squared and tilted ($t=0.5$) loss on test subset of \textit{corn} data. For three initializations the approximation fails to converge and $q$-optimal decisions have very high risk, but the decision-maker (blue) is able to correct for the failure.}
    \label{fig:hs}
\end{figure}

Variational approximations have trouble with multimodal or otherwise complicated posteriors. Yao et al.~\citet{Yao2018diditwork} use models with horseshoe priors \citep{carvalho2009} for demonstrating this,  and we use the same model family to demonstrate how we can still recover good decisions even when this happens.
We use the sparse regression model
\begin{align*}
& y | \beta \sim \mathcal{N}( X \beta, \sigma ), \quad  \beta_j | \tau, \lambda, c \sim \mathcal{N}(0, \tau^2 \tilde \lambda_j^2) ,\\
& \lambda_j \sim \mathrm{C}^{+} (0,1),   \quad   \tau   \sim \mathrm{C}^{+} (0,\tau_0), \quad c^2 \sim \mathrm{Inv\!\!-\!\!Gamma} (2,8),
\end{align*}
following closely a classification model proposed by Piironen and Vehtari~\citet{piironen2017hyperprior} with public Stan  code\footnote{{https://github.com/yao-yl/Evaluating-Variational-Inference/blob/master/R\_code/glm\_bernoulli\_rhs.stan}}, but change to Gaussian likelihood suitable for regression problems. We apply the model on the \textit{corn} data \cite{chen2009bayesian}, that we randomly split into equally sized training and test subsets.

For this model the quality of the variational approximation is sensitive to random initialization and the stochastic variation  during the optimization, so that occasionally the posterior is reasonable whereas for some runs it converges to a very bad solution. Figure~\ref{fig:hs} illustrates the risk for 10 independent runs with different random seeds, showing that for 7 runs all decision-making strategies are sufficient. However, for 3 of the runs the $q$-optimal decision fails miserably, yet the parametric decision-maker is able to recover essentially $p$-optimal decisions. In other words, we show that the proposed strategy is able to correct for the mistake in the posterior.

\subsection{Model Faithfulness}

The main goal of our work is to improve the decisions made given a specific probabilistic model, which means the decisions should remain faithful to the underlying model: The decision-maker should correct for mistakes in the posterior approximation but not in model miss-specification, since the users needs to be able to rely on interperations of the model. Next we show the regularization strategy proposed in Section~\ref{sec:decisionmaker} achieves this.

\begin{figure}[t]
    \centering
    \includegraphics[width=0.35\textwidth]{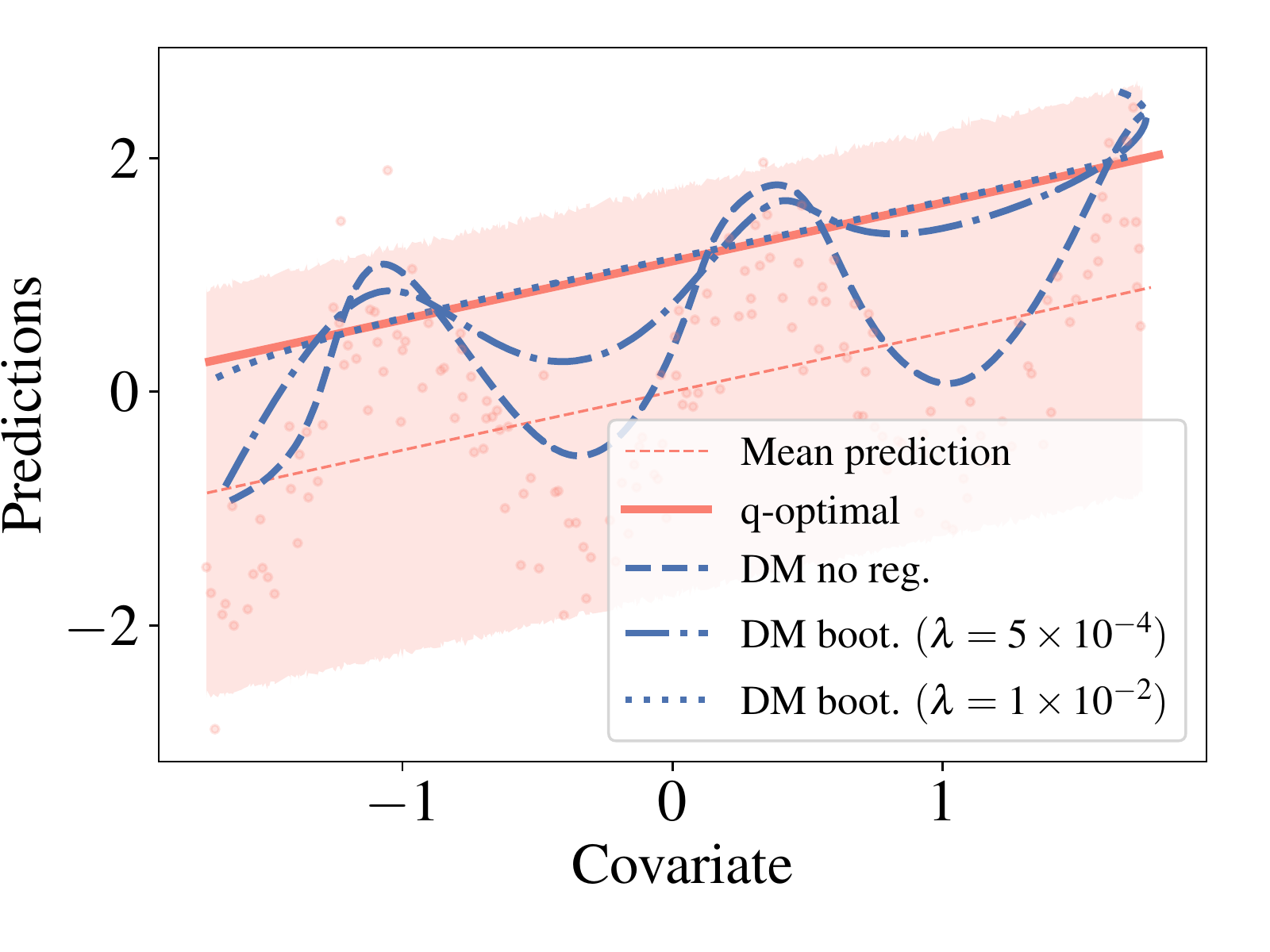}
    \caption{Without regularization the decision-maker can ignore the underlying linear model (red), but the proposed regularization strategy prevents this and provides predictions that are compatible with the model and only account for approximation error. Here the $q$-optimal decisions are also $p$-optimal, because there is no approximation error.}
    \label{fig:covariates}
\end{figure}

Figure~\ref{fig:covariates} shows the predictive distribution of intentionally incorrect model, a linear model for highly non-linear data. Even though the decision-maker only takes as input the predictive distribution (red shaded area), a flexible enough DM learns to ignore the model and returns the upper quantile of the data distribution that minimizes the risk. This can be done because there exists a mapping from the predictive quantiles back to the covariates, and hence the neural network can implicitly use the covariates themselves for predictions. In other words, it can learn to directly map the non-linear data, without conforming to the linear model.

Regularizing with \eqref{eq:h_simple_prior} smoothly transitions the decisions towards the $q$-optimal decision, here a linear function of the covariates. Sufficiently strong regularization prevents the flexible DM module from overriding the model.

\section{Conclusion}

Even though Bayesian decision theory \citep{berger} elegantly separates inference from decisions, this separation is constantly being misused by applying decision rules derived for the true posterior to approximations. This may dramatically affect the decisions, typically in a manner that is difficult to notice since there is no access to the true $p$-risk.

Some effort has been made to remedy this, based on directly modifying the inference process to account for the eventual decision loss \citep{lacoste2011approximate,cobb2018loss,lcvi,abbasnejad}, but these works  are computationally costly and have demonstrated only marginal improvements. We proposed the alternative of retaining the original approximation and changing the decision-making process instead, by replacing the analytic decision with parametric decision-making module. This has the significant advantage of being agnostic to the inference strategy being used, making the approach applicable to, e.g., improving predictive accuracy of models for which inference has already been carried out. To demonstrate the generality of our method, we ran several experiments on top of posterior approximations computed by off-the-shelf probabilistic programming system Stan \citep{carpenter2017stan}, demonstrating improved decisions for cases where the posterior approximation is inaccurate.

While the notion of directly optimizing for empirical risk may feel somewhat unorthodox in a strict Bayesian sense, we presented several technical arguments supporting the validity of the approach. By building on generalized Bayesian inference \citep{bissiri} we can reformulate reasoning over the decision-makers as justified Bayesian inference, and we provide a prior distribution that allows controlling for how faithful we remain to the underlying model, with automatic procedure for selecting parameters that does not allow significant deviation. While some theoretical questions of interest still remain, we have demonstrated that the quantitative value of predictive Bayesian models can be improved by incorporating flexible decision-makers borrowed from the deep learning literature without compromising the interpretability of Bayesian modeling.

\section*{Acknowledgements}
The work was supported by Academy of Finland, under grant 1313125, as well as the Finnish Center for Artificial Intelligence (FCAI), a Flagship of the Academy of Finland.
We also thank the Finnish Grid and Cloud Infrastructure (urn:nbn:fi:research-infras-2016072533) for providing computational resources.

\clearpage
\bibliography{refs}
\bibliographystyle{unsrt}
 
\end{document}